# Entity Aware Syntax Tree Based Data Augmentation for Natural Language Understanding

Jiaxing Xu[a,*], Jianbin Cui[b], Jiangneng Li[a], Wenge Rong[b] and Noboru Matsuda[c]

[a]*Nanyang Technological University, 50 Nanyang Ave, 639798, Singapore*
[b]*Beihang Uiversity, XueYuan Road No.37, Beijing, 100191, China*
[d]*North Carolina State University, Raleigh, 27695, NC, USA*

## ABSTRACT

Understanding the intention of the users and recognizing the semantic entities from their sentences, aka natural language understanding (NLU), is the upstream task of many natural language processing tasks. One of the main challenges is to collect a sufficient amount of annotated data to train a model. Existing research about text augmentation does not abundantly consider entity and thus performs badly for NLU task. To solve this problem, we propose a novel NLP data augmentation technique, Entity Aware Data Augmentation (EADA), which applies a tree structure, Entity Aware Syntax Tree (EAST), to represent sentences combined with attention on the entity. Our EADA technique automatically constructs an EAST from a small amount of annotated data, and then generates a large number of training instances for intent detection and slot filling. Experimental results on for datasets showed that the proposed technique significantly outperforms the existing data augmentation methods in terms of both accuracy and generalization ability.

## 1. Introduction

Natural language understanding (NLU) is one of the most fundamental components in natural language processing applications Goddeau et al. (1996). It aims to extract the intent and slots of the sentence Tur et al. (2011). Building an NLU system is a challenging task since it requires a large amount of application-specific data that are often hard to collect Bhagat et al. (2005).

With the recent development of deep learning techniques, most of the NLU approaches are based on neural networks Yao et al. (2013); Liu et al. (2015). The neural-network-based models normally need a large number of corpora **?**. It is a common belief that the more the training data, the better the performance of the resulted model. However, labeling a large amount of training data is very expensive. Therefore, the primary challenge is not only about collecting the corpus data, but also annotating the data with the type of intents. To overcome this challenge, data augmentation has been attached much attention Wei and Zou (2019).

Proper data augmentation is an alternative solution for improving a system's overall performance. In addition, data augmentation can also help the model converge faster, prevent overfitting, and avoid falling into a local optimum. Earlier approaches used rule-based methods, i.e., regular expressions Kuhn and De Mori (1995). However, it is difficult for a finite set of manually created regular expressions to cover all possible expressions. In addition, when performing named entity resolution, regular expressions can hardly generalize beyond the predefined list of possible words Aliod et al. (2006). As such it is challenging indeed to build their own NLU models with higher performance at a lower cost.

With the development of machine learning technique, many advanced approaches have been proposed, e.g., word substitution Zhang et al. (2015), back translation Xie et al. (2019), text surface transformation Coulombe (2018), and random noise injection Xie et al. (2017). However, these methods are not specifically designed for NLU task. Hence, additional noise may be introduced in the process of data augmentation.

Since syntax tree can better represent semantic information and sentence structure, some researchers have also tried to apply syntax tree to text data augmentation. For example, Sahin and Steedman (2018) used a dependency tree to augment text. The dependency tree is capable of obtaining the long-range dependency relation between words, especially for long sentences. However, its parsing depends on the entire sentence and it cannot take the diversity of

---

*Corresponding author

jiaxing.xu@ntu.edu.sg (J. Xu)

ORCID(s): 0000-0003-2498-5812 (J. Xu)

partial phrase into account in the process of data augmentation Luo et al. (2019). Furthermore, the dependency tree can only be constructed for a specific sentence. Only this sentence is used as a reference when generating a new sentence. This makes entities, which is important for NLU tasks Ali (2020), not taken into consideration. It is then interesting to ask if entity and syntax tree can be combined to support data augmentation for an NLU task.

In this paper, we propose Entity Aware Data Augmentation (EADA), an innovative technique to generate an Entity Aware Syntax Tree (EAST) that is used to automatically create a large volume of training data for an NLU task. Entity awareness means that we rely on the entities in the sentence to build a syntax tree for text generation. Through data pre-processing, main entity determination, EAST growth and sub-tree combination, our framework can automatically construct an entity aware syntax tree from a small scale of annotated data. Afterwards, the Mock Sentence Generator (MSG) generates a large number of texts to train a neural-network based model for intent detection and slot filling. The experiments on SNIPS, ATIS and CoNLL-2003 datasets prove EADA's potential.

The main contributions of this paper include the following:

1  We propose a new type of syntax tree (EAST) that can efficiently represent sentence with their entities.

2  We further put forward a data augment method called EADA that can automatically construct an EAST to represent a linguistic structure of the training data from a small set of actual training data. EAST is then used to generate a large number of mock sentences for training a target NLU model.

3  We provide empirical evidence showing that our method outperformed many state-of-the-art methods and commercial NLU systems.

4  We demonstrate the potential of EADA to facilitate the creation of NLU models via the task-specific data augmentation.

The rest of this paper is organized as follows. Section 2 discusses existing work from three aspects: natural language understanding models, data augmentation in natural language processing and syntax tree. Section 3 introduces the problem we want to tackle. Our proposed methods will be presented in detail in Section 4, followed by experiments in Section 5.

## 2. Related Work

This Section aims at contextualizing the proposed data augmentation method for natural language understanding by introducing some relevant works available in the literature. First, the existing works of NLU will be introduced; then, the application of data augmentation methodologies to natural language processing will be treated; last, some kind of syntax tree used to represent sentences will be discussed.

### 2.1. Natural Language Understanding

Natural language understanding (NLU) is a key component of a task-oriented dialogue system, and its quality has a significant impact on the overall quality of the dialogue system (Li et al., 2017). NLU aims to map a given query onto a pre-defined set of semantic slots. Usually, there are two types of representations (Chen et al., 2017). One is discourse-level categorization, e.g., intent detection and discourse categorization. Intent detection is to map a given query onto one of the pre-defined intents. The other one is word-level information extraction, e.g., named entity recognition (Lample et al., 2016) and slot filling (Mesnil et al., 2014). Unlike the discourse-level categorization, word-level information extraction is usually defined as a problem of sequence labeling. A general approach is to annotate individual words in a given query with the pre-defined semantic labels. An ideal feature of NLU is its versatility, the capability of switching a domain with a small amount of data without losing the knowledge learned from a previous domain.(Ni et al., 2020) The versatility becomes critical for a practical application of NLP, in particular when access to large training data is limited (Gasic et al., 2015; Wang et al., 2015; Ilievski et al., 2018).

Gasic et al.(2015) and Wang et al. (2015) have committed to creating a domain-independent, general dialogue policy. Ilievski et al. (2018) introduce a transfer learning method to mitigate the effects of the low in-domain data availability. All of the above methods tried to obtain more information from a small amount of training data. As such in this research we proposed an EADA to generate an entity aware syntax tree (EAST) as a template from a small amount data to generate a large text corpus for training a neural-based NLU model. EADA takes entities into consideration for represent NLU dataset with less loss, which can generate better sentences for augmentation.

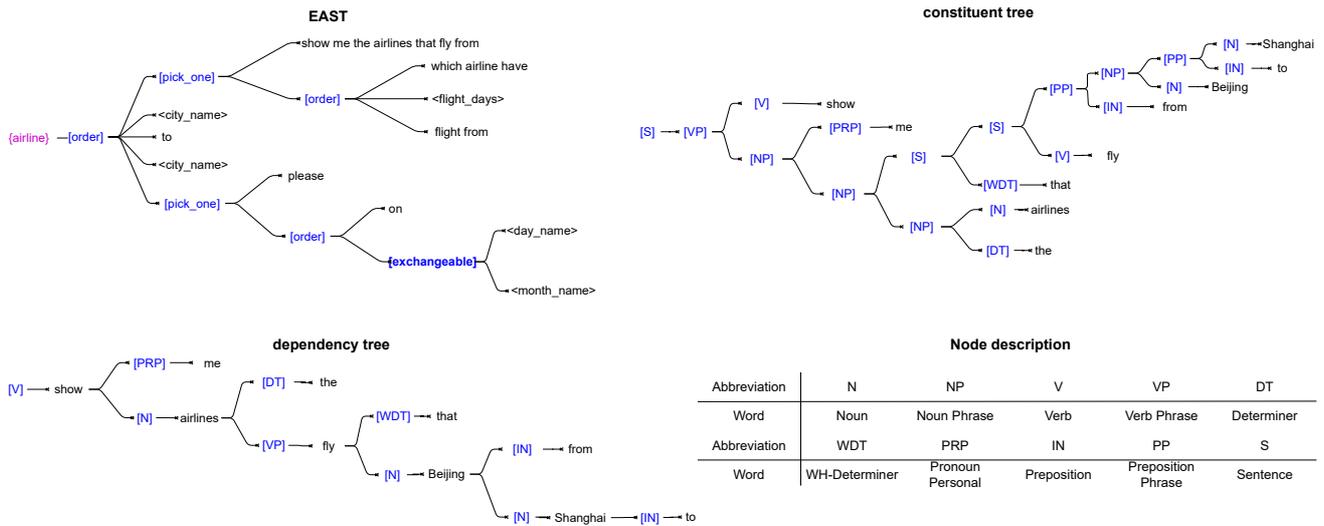

**Figure 1:** Different syntax tree of sentence "show me the airlines that fly from Beijing to Shanghai"

## 2.2. Data Augmentation in Natural Language Processing

Data augmentation has long been recognized as one essential mechanism for preparing a larger amount of training data. It has been proven effective in computer vision (Cubuk et al., 2019; Zoph et al., 2020). However, in the field of NLP, it is difficult to augment texts due to the high complexity in representation and context dependence. The augmented text must be not only grammatically sound but also meaningful. This kind of approach includes replacing randomly selected words from a text with a synonym in a given thesaurus (Zhang et al., 2015), using pre-trained GloVe embedding to find nearest neighboring words (Jiao et al., 2020), replacing words that have low TF-IDF scores (Xie et al., 2019), or predicting words in a text that are better to be replaced (Kobayashi, 2018). Easy Data Augmentation (EDA) is the best performing method of this type, which has four components including synonym replacement, random insertion, random swap, and random deletion (Wei and Zou, 2019).

Other commonly used data augmentation methods include back translation (Xie et al., 2019), text surface transformation (Coulombe, 2018), random noise injection (Xie et al., 2017), instance crossover augmentation (Luque, 2019), generating sentences with the opposite voice (e.g. from active voice to passive voice and vice versa) according to the dependency tree of the sentence (Coulombe, 2018).

Moreover, recently many neural network based text augmentation methods use generative models to fine-tune training data (Hu et al., 2017; Yoo et al., 2019; Kumar et al., 2020; Shin et al., 2019; Hou et al., 2021; Chang et al., 2021). Data Augmentationn with a Generation Approach (DAGA) is an augmentation method for NLU with language models trained on the linearized labeled sentences (Ding et al., 2020). TextFooler introduces adversarial attacks to select the most important word to replace for augmented sentence generation (Jin et al., 2020). The neural-network based models require a large amount of training data. Our proposed method, on the other hand, does not invoke a neural network, and only requires a small amount of data to augment text data.

Common text augmentation may not work well for all NLP tasks (Feng et al., 2021). In other word, different tasks need data augmentation method adapted by referring their own trait. The proposed Entity Aware Data Augmentation (EADA) method is a lightweight and efficient data augmentation framework which was designed for NLU tasks. The entity information in the sentence is fully considered, and the common collocations appearing in the sentence are arranged and combined on this basis. In addition, in the process of data augmentation, no other external information is introduced except the original dataset.

## 2.3. Syntax Tree

To represent sentences as a tree, there are many different types of syntax trees that can annotate words with a semantic label. (Fromkin et al., 2018) The constituent tree and the dependency tree are two classical syntax trees. The constituent tree represents a syntactic structure of a sentence in the context-free grammar. The leaf nodes of the constituent tree are associated with the words in the input sentence while other intermediate nodes are part of the marked phrase that represents the constituents of the sentence. On the contrary, the dependency tree represent dependency among words in a sentence in particular grammatical components such as "subject-predicate-object" and "attribute-adverbial-complement". Each node in the dependency tree is a word and each link is a dependency. The dependency tree does not take phrasal constituents into account.

**Table 1**
An illustrative example of natural language understanding.

| How's | the | weather | in | Beijing | today |
|-------|-----|---------|-----|---------|-------|
| O | O | O | O | LOC | DATE |
| Ask weather | | | | | |

These syntax trees can be used to generate sentences. For example, Sahin and Steedman (2018) augmented text by removing dependency links from the dependency tree, and moving the tree fragments around the root. The existing syntax tree labels a sentence with general grammatical labels, which means that we need to introduce additional annotations or prior knowledge to construct a syntax tree. Our Entity Aware Syntax Tree (EAST) can just use the label in the training data of NLU task without extra information. A simple example is shown as Figure 1. Different from the constituent tree and dependency tree, in EAST, only leaf nodes represent words, and other nodes represent the semantic information in the sentence.

## 3. Problem Definition

Entity Aware Data Augmentation (EADA) aims to assist two major NLU tasks: 1) *Intent detection* to identify an intent of a given query sentence, and 2) *slot filling* to convert the query into a domain-specific template. An intent is the category of the sentence that could be considered as the purpose of the speaker. A slot means the name entity type of the word, which is a real-world object, such as a person, location, organization, product, etc.

For example, for an utterance "How's the weather in Beijing today?", the intent detection model will first identify that the query is about the weather. The slot identifier will then tag each word with the intent-specific slot types, i.e., a location (LOC), a date (DATE), and other (O). The slots are tagged using the popular inside, outside, beginning (IOB) format, as shown in Table 1.

Assume that $\vec{x}$ is a sequence of $n$ words $w_i$ representing the given query $\vec{x} = w_1, w_2, ..., w_n$.

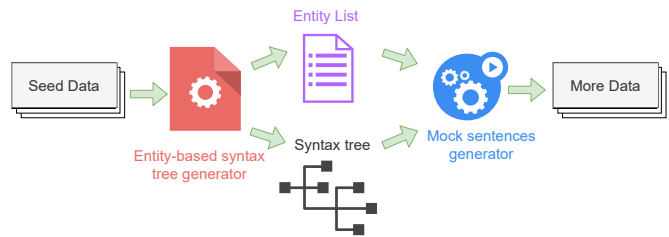

**Figure 2:** Framework architecture

*Definition 1. Intent Detection.* The task of intent detection is treated as a short-text classification task. Given a query sentence $\vec{x}$, intent detection is to compute its intent $\vec{y_i} = intent(\vec{x})$ where $\vec{y}$ is one of the pre-defined candidate classes.

*Definition 2. Slot Filling.* The task of slot filling is treated as a sequence annotation task. Given a query sentence $\vec{x}$, slot filling is to compute its associated slots $\vec{y_s} = s_1, s_2, ..., s_n$, where $s_i$ is the slot of $w_i$.

## 4. Framework

The entities in a sentence have a vital influence on the slot and intention of the sentence. Therefore, we should give priority to them when constructing the syntax tree. The proposed EADA framework has two major components:

1) an entity aware syntax tree generator, and 2) a mock sentences generator. The overall architecture of the proposed framework is shown in Fig. 2.

The proposed Entity Aware Syntax Tree Generator (EASTG) automatically generates an entity aware syntax tree (EAST) from the small amount of seed data. The EAST defines rules for synthesizing text data as shown in section 4.1. The detail of EASTG will be discussed in section 4.2.

The Mock Sentences Generator (MSG) then generates a massive amount of training data with the help of the EAST. The augmented training data will be then annotated with slots for each intent. For training on small amounts of samples, there is often a problem of overfitting, where sentences with a certain word are easily classified into a specific class during intent detection. MSG is proposed to solve this problem. Since EAST can represent different sentence structures, the sentences it produces are not only more diverse but also contain more information (Details of MSG are provided in section 4.3).

Since the generated sentence emphasizes the entity features, we hypothesize that an NLU model trained with EAST augmented data will perform better than just using original data or generating imitation sentences without pay attention on entities.

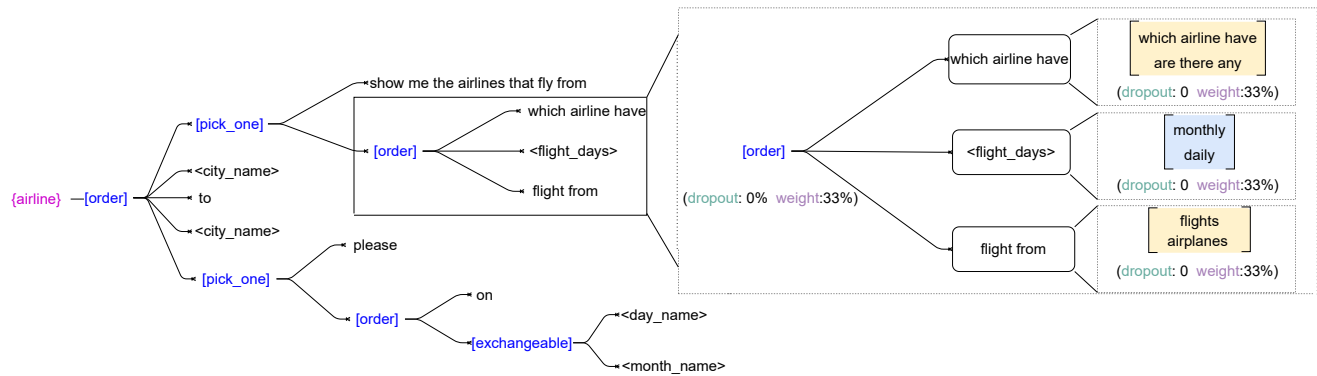

**Figure 3:** Entity aware Syntax Tree for Intent "airline". Blue nodes are control nodes and others are content nodes.

## 4.1. Entity Aware Syntax Tree

A common technique to build a rule based framework for data augmentation is to use regular expressions. However, to represent a rich linguistic structure of training data, regular expressions are not very applicable. In particular, due to the diversity of entities and their contexts, traditional regular expressions cannot handle the replacement of a large corpus well. Therefore, we propose the entity aware syntax tree (EAST) that is essentially a super-set of traditional regular expressions. The EAST represents an AND/OR structure of queries/sentences for a particular intent. The lexical information of the query is abstracted to be replaced with words in a given dictionary. Each intent has one and only one EAST. An example of an EAST for an intent "airline" is shown in Fig. 3. The definition of an EAST is shown as follows:

1. An EAST is an ordered tree.

2. A root node represents the query's intent.

3. An EAST contains content nodes, which, by definition, are leaf nodes, and all leaf nodes are content nodes. The content nodes are used to generate a specific content of a sentence. There are two kinds of content nodes:

- **Fixed content nodes**, which generate tokens from a predefined dictionary.

- **Entity content nodes**, which generate tokens from a large lexicon while tagging the slot name when generating.

4. An EAST also contains control nodes. Those control nodes are non-leaf nodes in the tree. Control nodes are used to control the specific generation path of the sentence. The control nodes only present the structure of the sentence and will not contribute words to the generated sentences. There are three kinds of control nodes:

- **Order node** An order node is used to restrict the generator to generate each child node of the order node in order.

- **Pick-one node** A pick-one node is used to restrict the generator to generate a child node of the pick-one node.

**step1: Data preprocessing**

(1) which airlines have daily flights from denver to san francisco April 1st

(2) are there any monthly airplanes from boston to dallas 4th May

(3) show me the airlines that fly from Beijing to Shanghai please

which airlines have <flight_days> flights from <city_name> to <city_name> on <month_name> <day_number>

are there any <flight_days> airplanes from <city_name> to <city_name> on <day_number> <month_name>

show me the airlines that fly from <city_name> to <city_name> please

**step2: Determin the main entity**

| Entity | flihgt_days | city_name | month_name | day_number |
|---|---|---|---|---|
| Percentage | 66% | 100% | 66% | 66% |

Main_entity

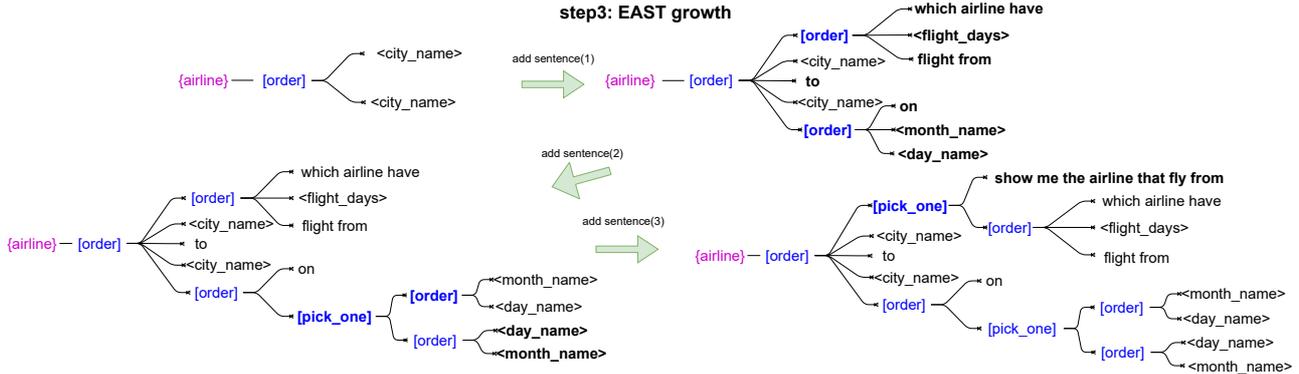

**step3: EAST growth**

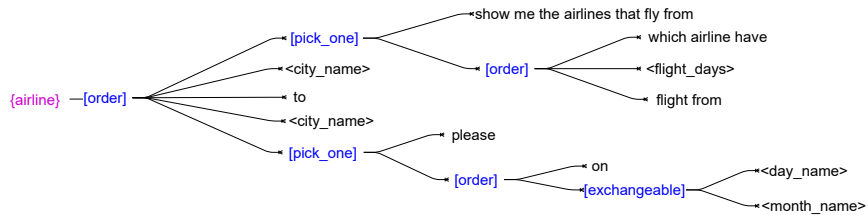

**step4: Sub-tree combination**

**Figure 4:** A simple process of generating an EAST

- **Exchangeable node** An exchangeable node is used to restrict the generator to generate all child node of the exchangeable node in any order.

5. Each node of the EAST contains a weight and an optional dropout value that allow the generator to discard some nodes to improve the generality.

## 4.2. EAST Generator

The following three steps are performed to construct an EAST from the training data automatically:

1. Determine the main entity. To construct an EAST for a given intent, the main entity of the intent needs to be identified. We will set a main entity threshold $t$ in (0, 1). The entity occurring more than $t$ in all entities will be treated as the main entity. If none of the entities occurring more than $t$, the most appearing entities will be chosen as the main entity. Once the main entity is determined, the EAST is initialized.

2. EAST growth. For each intent, we iterate all of its sentences and compare them with the current EAST. When the sentence's entities match with the tree, the content node is merged with the tree, and add the word to the dictionary of the content node. When the entities do not matched, a new branch is added to the tree as a pick-one node. The weight of the node is then added to ensure that every sentence provides an equal contribution. The method we used to add the node weight and node dropout is: if a branch is composed of multiple sentences, the weight of the node is obtained by dividing the number of times the corresponding node appears in these sentences by the total number of sentences. At the same time, the number of sentences with no content between the entity nodes is counted and divided by the total number of sentences to get the dropout of the nodes between the entity nodes.

3. Sub-tree combination. After generating all the subtrees, we use the order node to connect them into a tree. In addition, the existence of interchangeable leaf nodes is tested. Two leaf nodes are exchangeable when they are

---

**Algorithm 1** Process of construct EAST

---

**Input**: The Training Dataset $D = (S_i)_{i=1}^{N}$, where $S_i = (I, (w_j, s_j)_{j=1}^{m_i})$ is an annotated sentence with intent $I$ and $(w_j, s_j)$ is a word and its slot in the sentence. Main entity threshold $t$;

**Output**: An entity aware syntax tree $EAST$;

1:   Replace the entity in each $S$ in $D$ with its label; Collect entity list for every label. Categorize sentences according to the intent;
2:   **for** $S$ in every intent **do**
3:     **if** there is some entities $E$ occurrences for more than $t$ sentences **then**
4:       $main\_entity \leftarrow E$
5:     **else**
6:       $main\_entity \leftarrow$ the entity with the most occurrences
7:     **end if**
8:   **end for**
9:   use $main\_entity$ as initial leaf nodes of $EAST$;
10:   **for** $S$ in $D$ **do**
11:     compare $S$ with $EAST$
12:     **if** Entity matches **then**
13:       insert/append content node
14:       set/update weight/dropout
15:     **else**
16:       add/append pickone node and subtree
17:       set/update weight/dropout
18:     **end if**
19:   **end for**
20:   link leaf nodes with ordered node, add an exchagable node if both AB and BA existing

---

instantiated with two words interchangeably (e.g. October 1st and 1st October). The interchangeable leaf nodes are represented using an exchangeable nodes.

The whole process is shown in Algorithm 1, and a simple example is shown in Fig. 4. Through the EAST generator, we can automatically convert an unstructured natural language dataset into a structured EAST.

## 4.3. Mock Sentences Generator

A root node of an EAST represents an intent, which implies that there will be several EASTs created each for a particular intent. A synthetic sentence will be generated by traversing the EAST from the root node to a leaf node as follows. Each nodes in the EAST will be executed by following its corresponding function. If the root node has multiple order nodes, one of them is randomly selected, and the following steps will be applied:

When a node is the order node, all its child nodes will be added to the sentence in order. When a node is the pick-one node, a child node will be randomly selected by referring its weight calculated before and added to the sentence. When a node is the content node, there will be two cases: 1) It is not an entity word. The content node usually contains many replaceable phrases with the same meaning. During generation, we randomly select a phrase based on the proportion of them in the sample training set. 2) It is an entity word. We first randomly select one from the entity list as the candidate entity. Then use the cosine similarity of word embedding to find its K nearest neighbors and choose one to fill in this position and add it to the generated sentence. The cosine similarity between the candidate and the entity word will also be considered as the weight. K equals 5 in the following experiments.

Since EADA only use the entity that were seen in the training dataset, all generated sentences are properly labeled. Since EADA does not do any manual control of the fluency of generated sentences, which means the generated sentences will not be 100% grammatical. Actually, there can be a trade-off between diversity and validity of augmented instances (Dai and Adel, 2020). We believe a small quantity of noise can help models be more robust.

---

**Table 2**
Dataset statistics.

| Dataset | SNIPS | ATIS |
|---|---|---|
| Vocab Size | 11,241 | 722 |
| Average Sentence Length | 9.05 | 11.28 |
| #Intents | 7 | 21 |
| #Slots | 72 | 120 |
| #Training Samples | 13,084 | 4,478 |
| #Validation Samples | 700 | 500 |
| #Test Samples | 700 | 893 |

## 4.4. Implementation Details

We use $t = 0.5$ as the main entity threshold. As for the NLU models, 1) Capsule-NLU: We use a batch size of 16 and learning rate of 0.001. We set embedding size to 1024 and epoch number to 60. 2) LSTM-NLU: We use a batch size of 128 and learning rate of 0.005. We set embedding size to 50, hidden size to 128 and epoch number to 100.

For each dataset, we split it into 8:1:1 for training, validation and test. For all experiments, we evaluate each model with the same random seed for 5-fold cross-validation. All the codes were implemented using PyTorch (Paszke et al., 2017). The experiments were conducted in a Linux server with Intel(R) Core(TM) i9-10940X CPU (3.30GHz), TITAN X GPU, and 64GB RAM.

## 5. Experiment

This section describes a performance evaluation of EADA on a series of real-world datasets. We introduce the datasets and NLU models used for the evaluation followed by the results.

### 5.1. Datasets

To evaluate the effectiveness of our proposed framework, we designed and conducted several experiments on two real-world NLU datasets: 1) SNIPS Natural Language Understanding benchmark (SNIPS-NLU) and 2) Airline Travel Information Systems (ATIS) dataset. The statistical information on the two datasets is shown in Table 2. We also do experiments on CoNLL-2003 for the Name Entity Recognition (NER) task.

*Snips in-house NLU benchmark* (Coucke et al., 2018) consists of more than 16,000 utterances with 7 intents. We follow the train-test split of the original experiment.

*ATIS* provides not only a large number of messages, but also their associated intents and slots that can be used in training a classifier and a slot filler.

*CoNLL-2003* is a classic language-independent named entity recognition (NER) task datasets. It includes 1,393 English and 909 German news articles and mainly concentrates on four types of named entities: persons, locations, organizations and names of miscellaneous entities.

### 5.2. Data augmentation methods

We used three existing data augmentation methods, Word Substitution (WS), EDA (Wei and Zou, 2019) and Dependency Tree Morphing (DTM), against which EADA was compared. We designed a WS augmenter by combining a word embeddings augmenter, a TF-IDF augmenter, a contextual word embeddings augmenter, and a synonym augmenter. The augment approach is randomly chosen from these four augmenters when augmenting every sentence. The actual implementation is called nlpaug[1]. WS and EDA augmenter commonly use pre-trained models such as GloVe (Pennington et al., 2014), BERT (Devlin et al., 2019), and wordnet(Miller, 1998), which means they will introduce additional information. DTM (Sahin and Steedman, 2018) achieves the best performance among all published dependency tree based methods that have released source code. We used its official codebase[2] for our experiments.

---

[1] https://github.com/makcedward/nlpaug
[2] https://github.com/gozdesahin/crop-rotate-augment

**Table 3**
Slot filling and intent detection results on ATIS. Shown is average performance across ten independent runs. Bold shows that the increase has a statistical significance. Underline shows performance decreased after DA. There is an overall difference between the experimental results using and not using our method, (a) $t(5) = 4.58, p < .01.$ (b) $t(5) = 3.81, p < .01.$ (c) $t(6) = 4.49, p < .01.$ (d) $t(5) = 18.50, p < .001.$

| Model | DA Method | ATIS | | Snips | |
|---|---|---|---|---|---|
| | | intent(Acc) | slot(F1) | intent(Acc) | slot(F1) |
| Capsule-NLU | - | 0.950 | 0.952 | 0.973 | 0.918 |
| | WS | 0.954 | <u>0.895</u> | 0.976 | <u>0.891</u> |
| | EDA | 0.970 | <u>0.808</u> | 0.974 | <u>0.523</u> |
| | DTM | 0.956 | 0.960 | 0.974 | 0.919 |
| | EADA | 0.970 | 0.977 | 0.977 | 0.921 |
| LSTM-NLU | - | 0.958 | $0.978^b$ | 0.965 | $0.906^d$ |
| | WS | 0.964 | <u>0.904</u> | 0.970 | <u>0.871</u> |
| | EDA | 0.974 | <u>0.892</u> | 0.972 | <u>0.605</u> |
| | DTM | 0.969 | 0.979 | 0.968 | 0.920 |
| | EADA | 0.977 | $\mathbf{0.981}^b$ | 0.974 | $\mathbf{0.952}^d$ |
| SF-ID | - | $0.971^a$ | 0.954 | 0.964 | 0.917 |
| | WS | 0.973 | <u>0.825</u> | 0.920 | <u>0.820</u> |
| | EDA | 0.978 | <u>0.831</u> | 0.933 | <u>0.732</u> |
| | DTM | 0.973 | 0.969 | 0.970 | 0.919 |
| | EADA | $\mathbf{0.981}^a$ | 0.977 | 0.981 | 0.919 |
| SP+Bert | - | 0.968 | 0.952 | $0.976^c$ | 0.930 |
| | WS | 0.974 | 0.962 | 0.981 | 0.932 |
| | EDA | 0.974 | 0.968 | 0.982 | 0.934 |
| | DTM | 0.970 | 0.966 | 0.977 | 0.934 |
| | EADA | 0.978 | 0.978 | $\mathbf{0.984}^c$ | 0.941 |

**Table 4**
LSTM-NLU+EADA outperformed 5 existing NLU technologies in computing slot filling on SNIPS dataset measured by F1. Bold shows the best performance and underline shows second-best performance. Shown is average performance across ten independent runs.

| | train size=70 | | | train size=2000 | | |
|---|---|---|---|---|---|---|
| NLU provider | precision | recall | F1-score | precision | recall | F1-score |
| Luis | 0.909 | 0.537 | 0.691 | 0.954 | 0.917 | <u>0.932</u> |
| Wit | 0.838 | 0.561 | 0.725 | 0.877 | 0.807 | 0.826 |
| DialogFlow | 0.770 | 0.654 | 0.704 | 0.905 | 0.881 | 0.884 |
| Alexa | 0.680 | 0.495 | 0.564 | 0.720 | 0.592 | 0.641 |
| Snips | 0.795 | 0.769 | **0.790** | 0.946 | 0.921 | 0.930 |
| LSTM-NLU+EADA | 0.760 | 0.764 | <u>0.761</u> | 0.937 | 0.940 | **0.938** |

## 5.3. Baseline

We use several existing neural NLU models to verify the effectiveness of the proposed data augmentation method: 1) SF-ID (Haihong et al., 2019) designs a new iteration mechanism to enhance the bi-directional interrelated connections between intent detection and slot filling. 2) Stack-Propagation (Qin et al., 2019) can capture the intent semantic knowledge by directly using the intent information as input for slot filling. 3) Capsule-NLU (Zhang et al., 2019) propose a capsule based neural network model that accomplishes slot filling and intent detection via dynamic routing-by-agreement schema. Capsule-NLU reaches the highest performance comparing with the existing model of NLU. 4) LSTM-NLU independently deal with the two sub-tasks of NLU by simply using an LSTM (Ravuri and Stolcke, 2015, 2016) for intent detection and a BiLSTM-CNN-CRF (Ma and Hovy, 2016) for slot filling.

As for the NLU models, 1) Capsule-NLU: We use a batch size of 16 and a learning rate of 0.001. We set the embedding size to 1024 and the epoch number to 60. 2) LSTM-NLU: We use a batch size of 128 and a learning rate of 0.005. We set embedding size to 50, hidden size to 128 and epoch number to 100. Other models' settings are the same as their papers.

**Table 5**
Slot filling results on Conll-2003 dataset. All data augmentation methods generate extra double data. Shown is average performance across ten independent runs.

|            | precision | recall | F1    |
|------------|-----------|--------|-------|
| Bi-LSTM-CRF | 0.892     | 0.888  | 0.890 |
| +WS        | 0.930     | 0.944  | 0.937 |
| +DTM       | 0.930     | 0.947  | 0.939 |
| +EADA      | **0.941** | **0.953** | **0.947** |

Many commercial NLU services support training NLU models from only a few examples. We followed Coucke et al. (2018) and compared the best performing NLU model with EADA with existing commercial NLU services, including Microsoft's Luis[3], Facebook's Wit[4], Google's DialogFlow[5], Amazon's Alexa[6] and Snips(Coucke et al., 2018).

### 5.4. Model Validation Procedures

We mainly compared Capsule-NLU and LSTM-NLU crossed with four data augmentation methods (WS vs. EDA vs. DTM vs. EADA, in addition to no data augmentation) using the ATIS dataset and SNIPS dataset.

We then compared the best performing NLU model from the initial comparison against the existing commercial NLU services only for slot filling since they have not published their result of intent classification. Since we could not re-train the commercial NLU services, we did not apply any data augmentation for them. For each model crossed with data augmentation, we ran 5-fold cross-validation. The results shown below are based on the average among the five folds. Finally, we examined how the volume of augmented training data affects the model performance.

### 5.5. Main Results for NLU

The results on the ATIS and SNIPS dataset are reported in Table 3. The data show that EADA outperformed all other data augmentation methods across four NLU models for both slot filling and intent detection. However, we found that not all data augment method could improve the performance. For slot filling, LSTM-NLU with word substitution (0.904) performed slightly worse than without data augmentation (0.978). It can tell us that general text augment may not suit every NLP task and it is necessary to design task-specific data augmentation. EADA performs better because it can integrate the entities and sentence patterns of different intents to generate data that allows the model to learn more features. In other words, EADA is more suitable for targeted data augmentation for NLU tasks.

Table 4 shows the results of comparing the best-performing model (LSTM-NLU) with EADA against commercial NLU services on slot filling. The result shows that the recall and F1 scores of LSTM-NLU with EADA are higher than the target commercial NLU models on small-scale data (only 70 and 2000 samples for training).

It also shows that the commercial NLU services tend to have better precision scores. There might be several factors that could attribute the high precision, e.g. the nature of the training data (e.g., the class imbalance problem), simply the amount of training data, learning regularization, and etc. To build a practical NLP system, identifying more entities correctly allows the next module in the NLP pipeline to utilize more information. We would therefore argue that obtaining higher recall is more crucial than higher precision depending on the purpose of the NLP system to be built. EADA will significantly improve the performance of NLP system when higher recall in slot filling is essential.

### 5.6. Experiment for Name Entity Reconision

EADA only relies on the annotation of entity. In other words, intent labeling is not necessary for our proposed method and we can apply EADA only to NER tasks. Therefore, we conducted experiments on NER's common data set Conll-2003. The scale of this data set is larger, and the topic is not limited to a single domain. We choose Bi-LSTM-CRF (Lample et al., 2016), which is a classical NER model, as the base model. As shown in Table 5, our method outperforms both two commonly used text augmentation technologies.

---

[3] https://www.luis.ai
[4] https://wit.ai/
[5] https://dialogflow.com
[6] https://aws.amazon.com/lex/

**Table 6**
Ablation study on ATIS and SNIPS dataset. Shown is average performance across five independent runs.

| Module | | | ATIS | | Snips | |
|---|---|---|---|---|---|---|
| +regex | +EAST | +word_emb | intent(Acc) | slot(F1) | intent(Acc) | slot(F1) |
| | | | 0.958 | 0.978 | 0.965 | 0.906 |
| ✓ | | | 0.961 | 0.979 | 0.969 | 0.931 |
| ✓ | ✓ | | 0.968 | 0.979 | 0.970 | 0.944 |
| ✓ | ✓ | ✓ | **0.977** | **0.981** | **0.974** | **0.952** |

### 5.7. Ablation Study

We perform an ablation study to evaluate the importance of each component in EADA framework, i.e., the regex, EAST and word_emb. As we mentioned in the section of EAST that EAST is essentially a super-set of traditional regular expressions. So +regex means we only use the set of regular expressions without any tree structure. Furthermore, +EAST means with our tree structure but no word emb. As for +word_emb, we mentioned how we use the cosine similarity of word embedding to select candidates at the end of Mock Sentences Generator section. The base model is LSTM-NLU. Results show that all components can benefit the performance of NLU model (Table 6).

### 5.8. Volume of Training Data

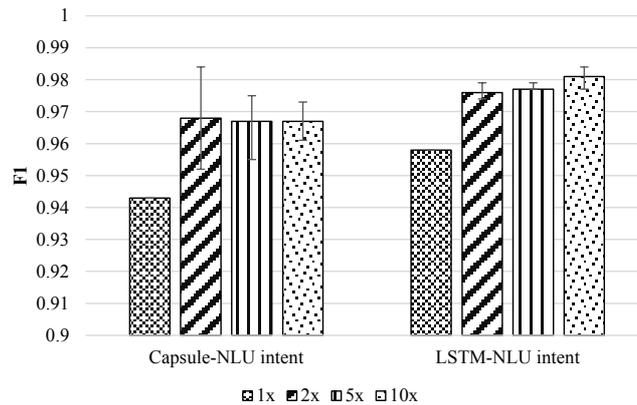

**Figure 5:** Intent detection results using EADA on ATIS dataset with different generated data. Black bars indicate the standard deviation.

We hypothesized that the more data there is, the better the model performance would be. To test this hypothesis, we compared the performance of capsule-NLU and LSTM-NLU with 2, 5, and 10 times data augmentation while all other hyperparameters remain unchanged. Fig. 5 shows an average accuracy score on intent detection aggregated across 10 trials, while Fig. 6 shows an average F1 score on slot filling aggregated across 10 trials. The results show that the performance of the model using EADA is better than the model without data augmentation. At the same time, we can find that different models require different augmentation factors to achieve an optimal performance.

### 5.9. Case Study

Here we further use an example to show the comparison of four augmentation methods (EADA, EDA, DTM, and WS) in Fig. 7. It is found that the sentences generated by EADA rely less on the original sentences. In other words, EADA's expanded data set is more diverse and contains more information. Moreover, due to the existence of dropout and the random selection of replacement words, the generated sentences may not conform to the grammatical rules. This situation exists in both our method and the other existing methods. It is normal and can be allowed because the existence of these noises enhances the robustness and generalization ability of the model.

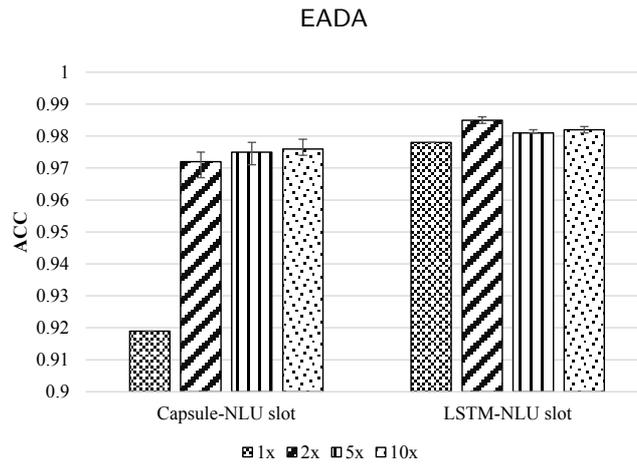

**Figure 6:** Slot filling results using EADA on ATIS dataset with different generated data. Black bars indicate the standard deviation.

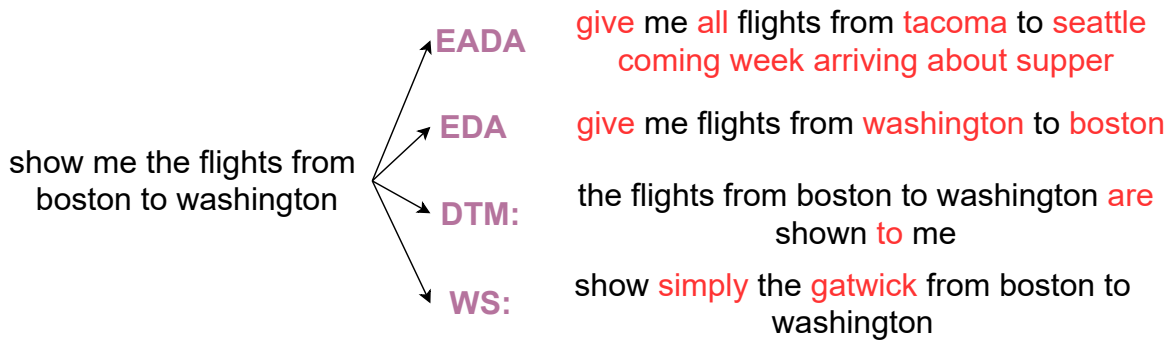

**Figure 7:** Comparison with three augmentation methods. Red words are words that did not appear in the original sentence.

**Table 7**
Intent statistics of the AMiner dataset.

| Intent | Sentences Number |
| --- | --- |
| Search scholar | 50 |
| Search publication | 50 |
| Chitchat | 116 |
| Search event | 51 |
| Search trend | 52 |
| Search news | 50 |
| Compare keywords | 54 |
| Talent map | 50 |
| Search venue | 49 |

### 5.10. Human in the Loop

Sometimes developer even don't have limited annotated data. EADA can also handle this intractable situation by take human in the loop. Since EAST can directly been designed by human, a few rules designed by expert will help to cold-start neural networks.

In cold-start scenarios, the NLU model is obtained without any training data. Only a set of rules is available. We design an experiment to empirically prove that our framework can outperform vanilla matching method. This experiment was performed on the AMiner dataset. No training data was used. For each intent, we handcrafted several rules. For each type of slot, we used the lexicon corpora described in the previous subsection.

**Table 8**

Name entity statistics of the AMiner dataset.

| Intent | Sentences Number |
| --- | --- |
| Keyword | 386 |
| Organization | 47 |
| Person | 62 |
| Date | 39 |
| Venue | 40 |
| Location | 30 |

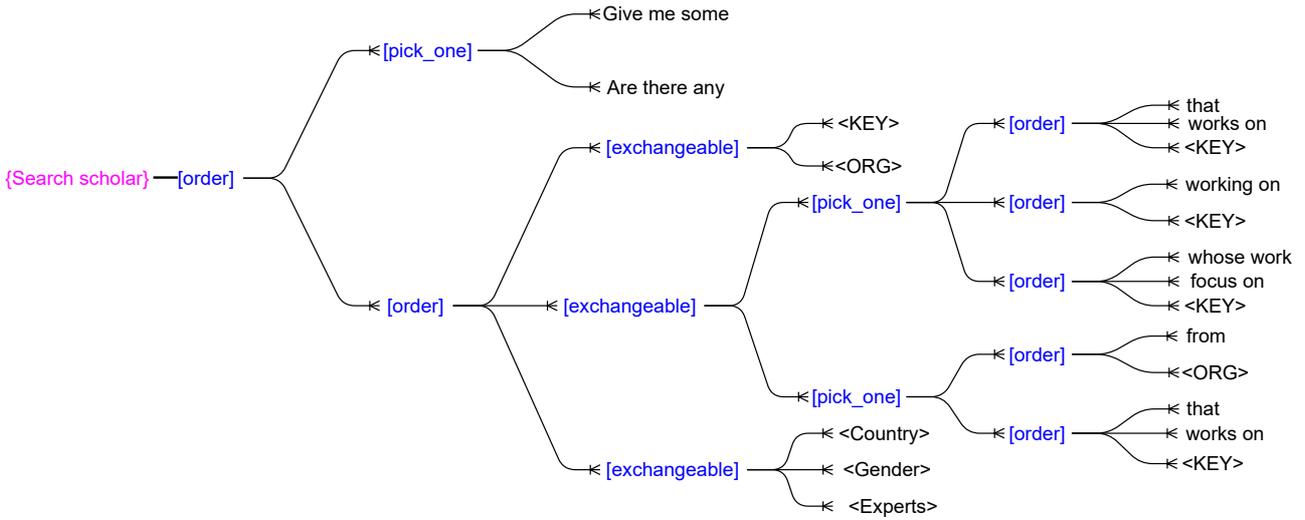

**Figure 8:** Entity Aware Syntax Tree for Intent "Search Scholar". Blue node is order node, red node is pick-one node, yellow node is exchangeable node and the green one is content node.

*AMiner* AMiner, the second version of ArnetMiner (Tang et al., 2008), is a free online academic search and mining system. We built an online dialogue system for AMiner using the proposed framework and constructed an internal dataset collected from interactions between real users and that dialogue system for evaluation. This dataset consists of both English and Chinese utterances. As there is a speech-to-text module for the dialogue system, some utterances were converted to text from speech, while others were typed in. The dataset contains among 600 utterances distributed 10 intents. Accompanying with this dataset is a lexicon corpus of named entities in AMiner, used to train the slot filling models. For each type of named entity, we only keep the words that occurred more than 200 times in AMiner. The intent statistics is as Table 7 shows and the named entity statistics is as Table 8 shows.

*Pre-training* For Chinese word embeddings, we used the the pre-trained embeddings provided by (Li et al., 2018), trained on the *Mixed-large* corpus. For English words, we used the pre-trained GloVe (Pennington et al., 2014) model. The pre-trained BERT model we used is the *BERT-Base, Multilingual Cased* model provided officially by Google [7] without fine tuning. But because of the complexity of this task-oriented dataset, the BERT method can't achieve better results than word embedding.

*Result* The evaluation metrics are the same as in (Tjong Kim Sang and De Meulder, 2003). Following that work, we used macro-F1 for the evaluation of both intent classification and slot filling. A slot is correct only when it is an exact match of both entity type and words. The EAST we got for the "search scholar" intention is shown as Figure 8. We compared our method with regular expressions generated from the same set of rules. The results are shown in Table 9 and Table 10. The results have shown that given the same input, our method could perform significantly better than regular expressions.

---

[7] https://github.com/google-research/bert

**Table 9**
Intent classification in a cold-start scenario

| Intent | LSTM-NLU+regex | | | LSTM-NLU+EADA | | | SP+Bert+EADA | | |
|---|---|---|---|---|---|---|---|---|---|
| | Precision | Recall | F1 | Precision | Recall | F1 | Precision | Recall | F1 |
| Search scholar | **0.941** | 0.640 | **0.762** | 0.593 | 0.640 | 0.343 | 0.615 | **0.940** | 0.503 |
| Search publication | 0.943 | 0.660 | 0.776 | 0.830 | 0.880 | 0.854 | **0.960** | 0.960 | **0.960** |
| Chitchat | 0.283 | **0.991** | 0.441 | 0.383 | 0.595 | **0.466** | 0.488 | 0.172 | 0.255 |
| Search event | 0.980 | 0.098 | 0.179 | **1.000** | 0.569 | 0.725 | **1.000** | 0.745 | **0.854** |
| Search trend | 0.973 | 0.462 | 0.632 | **1.000** | 0.692 | 0.818 | 0.980 | **0.923** | **0.951** |
| Search news | **1.000** | 0.280 | 0.438 | 1.000 | 0.660 | 0.795 | 0.833 | **1.000** | 0.909 |
| Compare keywords | 0.867 | 0.792 | 0.827 | 0.797 | 0.944 | 0.864 | **0.981** | **0.981** | **0.981** |
| Talent map | **1.000** | 0.140 | 0.246 | 0.900 | 0.720 | 0.800 | **1.000** | 0.980 | **0.990** |
| Search venue | 0.460 | 0.626 | 0.530 | **0.979** | **0.979** | **0.979** | 0.976 | 0.851 | 0.909 |
| Average | 0.717 | 0.427 | 0.535 | 0.750 | 0.768 | 0.759 | **0.796** | **0.861** | **0.827** |

**Table 10**
Slot filling in a cold-start scenario

| Method | Precision | Recall | F1 |
|---|---|---|---|
| LSTM-NLU+regex | 0.275 | 0.211 | 0.239 |
| LSTM-NLU+EADA | 0.404 | 0.425 | 0.414 |
| SP+Bert+EADA | **0.525** | **0.497** | **0.511** |

## 5.11. Theretical and Practical Implications

In general, existing generally-used text augmentation methods (eg. WS and EDA) may poisoning the NLU model. The existing tree-based text augmentation method (eg. DTM) can easily ingore the entity information in NLU dataset, and it also need extra prior knowledge for syntax tree construction. We proposed a new kind of syntax tree, EAST, which can naturally represent NLU dataset by aware of entity in it without any additional information. By using EAST, the EADA we presented in this paper is specifically designed for NLU task and can amazingly improve the performance of both RNN-based and Transformer-based NLU model. EADA outperforms with higher efficiency and generalization ability comparing not only with other 3 DA methods, but also 5 commercial NLU service. Moreover, EADA can be individually applied on sequence tagging task without rely on class label. The ablation study shows EADA benefits from three components. The case study indicates that sentences generated by EADA having more diversity with other text augmentation methods. Eventually we discuss the cold-start scenario of NLU task. We can take human in the loop to get a usable dataset from scratch, which is valuable when switch to a new domain. In the future, we will explore whether EADA can be extend to more NLP tasks.

## 6. Conclusion

In this research, we propose a novel syntax tree named EAST, which can well characterize the entities and intentions in the dataset. Based on EAST, we further put forward EADA, a template-based data augmentation method, to improve the NLU system. Experimental results verify the advantage of our method over not only other data augment methods, but also the state-of-the-art online NLU providers. Furthermore, we discussed the pertinence of data augmentation methods for NLU tasks, and explored the impact of the amount of generated data on the effect of neural networks.